\documentclass{article} 

\usepackage[accepted]{icml2025}

\usepackage{amsmath}
\usepackage{amsfonts}
\usepackage{bm}
\usepackage{url}
\usepackage{graphicx}
\usepackage{booktabs}
\usepackage{subcaption}
\usepackage{listings}
\usepackage{float}
\usepackage{amssymb}
\usepackage{tabularray}
\usepackage{hyperref}

\DeclareMathOperator*{\argmax}{arg\,max}

\newcommand{\phenombeta}{CA\nobreakdash-MAE-S/16\,}
\newcommand{\phenomone}{MAE\nobreakdash-L/8\,}
\newcommand{\phenomoneretrained}{MAE\nobreakdash-L/8\,}
\newcommand{\phenomtwo}{MAE\nobreakdash-G/8\,}
\newcommand{\dday}{Phenoprints\nobreakdash-16M\,}
\newcommand{\ddaynospace}{Phenoprints\nobreakdash-16M}
\begin{document}

\twocolumn[
\icmltitle{ViTally Consistent: Scaling Biological Representation Learning for Cell Microscopy}


\icmlsetsymbol{equal}{*}

\begin{icmlauthorlist}
\icmlauthor{Kian Kenyon-Dean}{rxrx}
\icmlauthor{Zitong Jerry Wang}{rxrx}
\icmlauthor{John Urbanik}{rxrx}
\icmlauthor{Konstantin Donhauser}{vl}
\icmlauthor{Jason Hartford}{vl,manchester}
\icmlauthor{Saber Saberian}{rxrx}
\icmlauthor{Nil Sahin}{rxrx}
\icmlauthor{Ihab Bendidi}{vl}
\icmlauthor{Safiye Celik}{rxrx}
\icmlauthor{Juan Sebasti\'{a}n Rodr\'{i}guez Vera}{rxrx}
\icmlauthor{Marta Fay}{rxrx}
\icmlauthor{Imran S Haque}{rxrx}
\icmlauthor{Oren Kraus}{rxrx}
\end{icmlauthorlist}

\icmlaffiliation{rxrx}{Recursion}
\icmlaffiliation{vl}{Valence Labs}
\icmlaffiliation{manchester}{University of Manchester}

\icmlcorrespondingauthor{Kian Kenyon-Dean}{kian.kd@recursion.com}
\icmlcorrespondingauthor{Oren Kraus}{oren.kraus@recursion.com}

\icmlkeywords{Machine Learning, ICML, MAE, drug discovery, microscopy, SSL, linear probing, biology, high-content screening, foundation models}

\vskip 0.3in
]

\date{}

\printAffiliationsAndNotice{}
\begin{abstract}

Deriving insights from experimentally generated datasets requires methods that can account for random and systematic measurement errors and remove them in order to accurately represent the underlying effects of the conditions being tested. Here we present a framework for pretraining on large-scale microscopy datasets that includes three steps: (1) curating a set of diverse and self-consistent training samples, 
(2) scaling 
training of an appropriate foundation model architecture on this dataset, (3) evaluating intermediate layers of the trained model to identify the best representation for downstream tasks. 
Using this strategy, we present the largest foundation model for cell microscopy data to our knowledge, a new 1.9 billion-parameter ViT-G/8 MAE trained on over 8 billion microscopy image crops.
Compared to a previous published ViT\nobreakdash-L/8 MAE, our new model achieves a 60\% improvement in linear separability of genetic perturbations and obtains the best overall performance on whole-genome relationship recall, batch correction replicate consistency, and compound-gene activity prediction benchmarks.\footnote{An earlier version of this work appeared at the NeurIPS 2024 Foundation Models for Science Workshop \citep{kenyon2024vitally}.}

\end{abstract}

\section{Introduction}

Microscopy images capture detailed information about a cell's responses to perturbations: indeed many early biological discoveries were made by biologists looking through microscopes studying how cells behave ~\citep{Hooke1665}. 
But with the advent of high throughput screening platforms that capture cells' responses to tens or hundreds of thousands of perturbations, we can no longer rely on people to study these images. Instead, we need representations of the cell images that capture the biologically meaningful information and capture the similarities and differences in cells’ response to perturbations.

\begin{figure}[t]
    \centering
    \includegraphics[width=\linewidth]{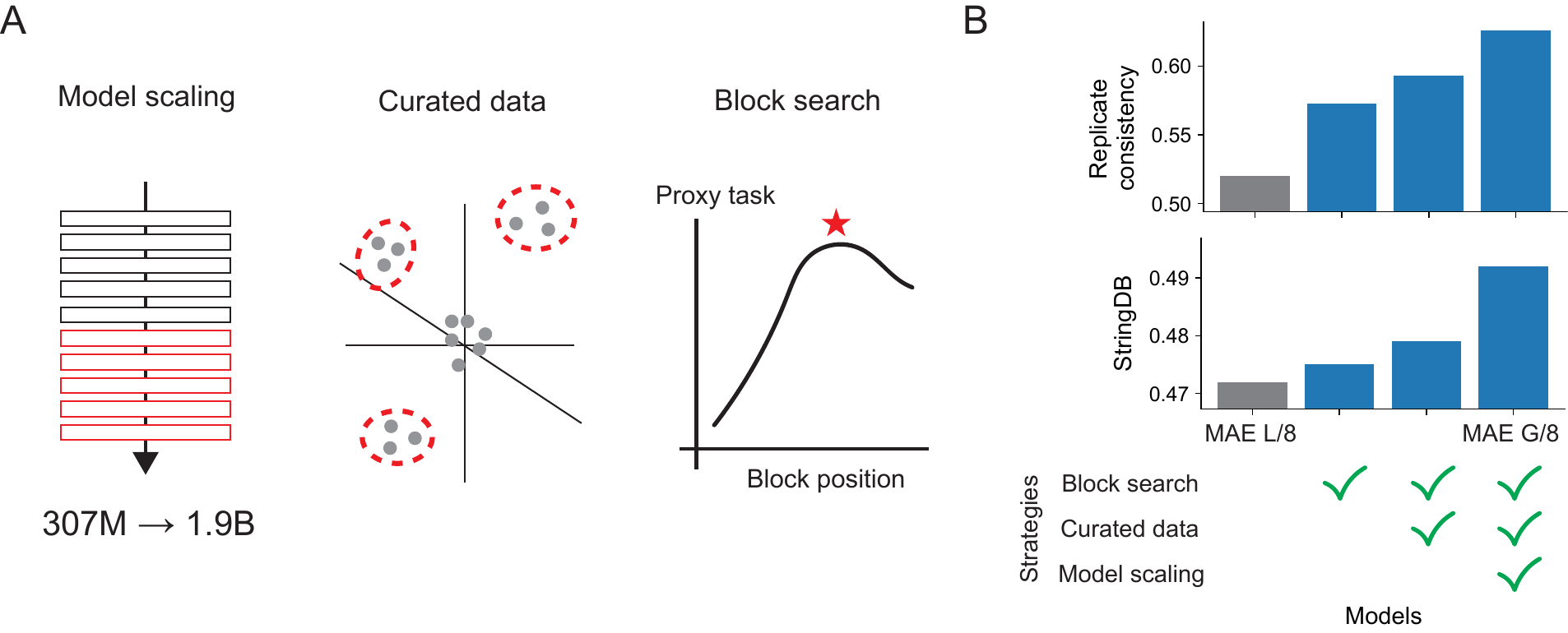}
    \caption{(A) Overview of performance gain from different MAE pretraining and inference strategies. (B) Example whole-genome results for \textbf{replicate consistency} and \textbf{biological relationship recall} on StringDB for models trained with different combinations of strategies, by model name and dataset (left to right):} 
    \label{fig:method}
\end{figure}
State-of-the-art (SOTA) deep learning methods for learning representations of microscopy leverage Vision Transformers (ViT)~\citep{dosovitskiy2020image} trained with self-supervised learning (SSL) techniques~\citep{SSL_cookbook} from large-scale screens~\citep{Doron2023UnbiasedSM,kim2023self,bourriez2024}. They have been shown to be very effective at representing the subtle changes in cell morphology in response to perturbations, and masked autoencoders \citep{he2022masked} in particular exhibit scaling laws for recovering known biological relationships~\citep{kraus2024masked}. However, given that this data has to be collected experimentally, there is a limit to how far we can scale these approaches before one runs out of data. To continue to improve these representations, we need to use experimental data as efficiently as possible.
In this paper, we show how curated training datasets combined with careful probing of model representations allows us to scale to larger models that significantly improve recall and separation of biological phenotypes. 

Careful curation of data has enabled better data efficiency in language models \citep{Llama3, eldan2023tinystories,javaheripi2023phi}. However, unlike language, we cannot simply rely on prior knowledge about the data source to weigh data (e.g. by up-weighting trusted sources like arXiv or wikipedia). Instead, we curate a training dataset that is relatively uniform with respect to the observed phenotypes.
Because many perturbations will not induce a morphological change, a random sample of images across all perturbations will be dominated by cells that look like unperturbed cells. We show that one can use weaker models to filter images to perturbations that induce a distinct morphological phenotype, thereby reducing the absolute training set size but increasing the relative diversity. The resulting dataset, \emph{Phenoprints-16M}, has $5\times$ fewer images than the dataset used to train \citeauthor{kraus2024masked}'s masked autoencoder, but we find that it leads to better performance when combined with scaled compute through longer self-supervised training.

Data curation gives one way of improving a model, but we can also optimize over the layer that we use to construct a representation. Traditionally, the finally hidden layer has been used \citep{Doron2023UnbiasedSM, kim2023self, kraus2024masked}, but there is some evidence from mechanistic interpretability \citep{templeton2024scaling} that earlier layers may learn more abstract features, which may provide more useful features for biological tasks.
We show that using a linear probing proxy task,
we are able to cheaply find the best performing intermediate block for many models. 
We find that using intermediate layers leads to better performance on downstream whole-genome relationship recall and perturbational replicate consistency at a lower computational inference cost, and that these results are consistent across a variety SSL ViTs trained on either microscopy or natural images.
Our proxy task involves fitting a set of {biological linear probing tasks} to evaluate representations learned by intermediate ViTs blocks for microscopy data (\S~\ref{sec:probe_results}). We find this task correlates extremely well with biological recall benchmarks which allows us to reliably probe the model's representations.

In summary, we present a series of data curation and model probing techniques that allow us to more effectively use experimental microscopy data.
By combining these insights, we are able to train a \textbf{new foundation model, MAE\nobreakdash-G/8}, a 1.9 billion parameter ViT-G/8 MAE trained on \dday for 48,000 H100 GPU hours on more than 8 billion samples drawn from the curated dataset (\autoref{fig:method}A, \S~\ref{sec:models}) resulting in significant improvements across a range of challenging biological benchmarks. These results indicate that the scaling properties first identified by \citet{kraus2023masked} extend to the {multi-billion parameter model regime} across a wide variety of newly examined biologically-motivated benchmark tasks. 

\section{Related work} \label{sec:related}

\begin{figure}[t]
    \centering
    \includegraphics[trim=0 30 0 0, clip, width=\linewidth]{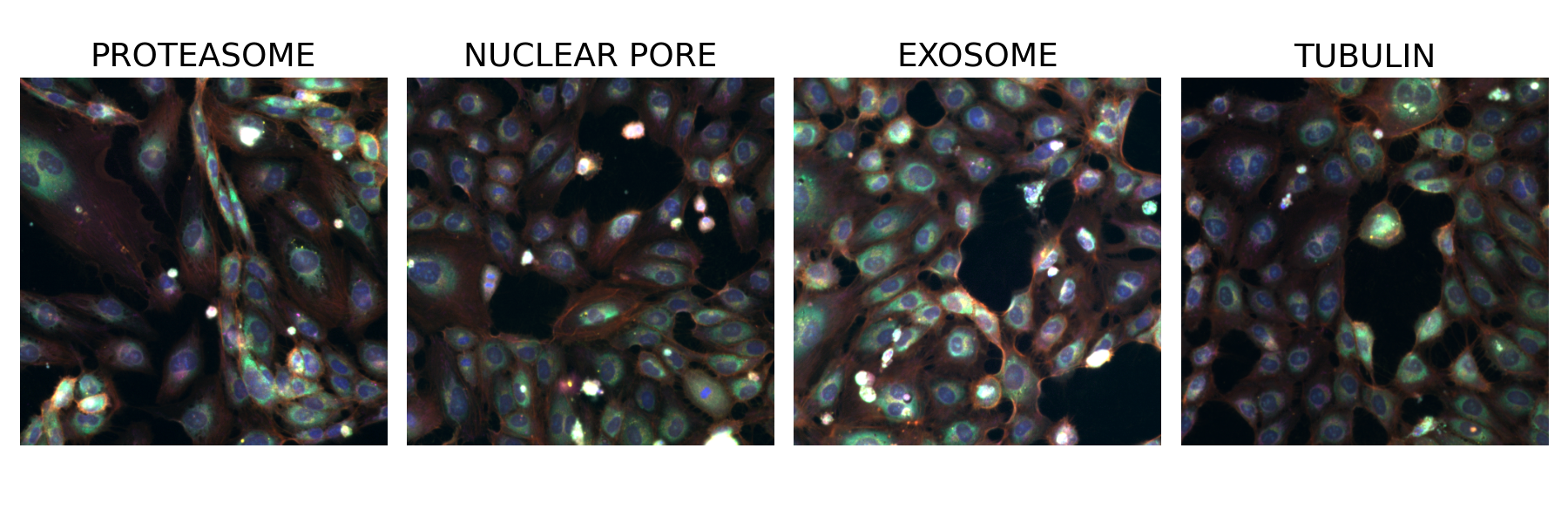}
    \caption{Samples for subset of groups in Anax 40-class functional gene group classification task.}
    \label{fig:anax_vis}
\end{figure}

\paragraph{Dataset Curation for Foundation Models.}
Dataset curation is crucial for enhancing the efficiency of foundation models, especially in large-scale contexts. Usual approaches to dataset construction are inspired by the image retrieval community~\citep{weinzaepfel2022learningsuperfeaturesimageretrieval, radenović2018finetuningcnnimageretrieval, berman2019multigrainunifiedimageembedding}. Existing methods often utilize pre-trained models for filtering and pruning, such as vision-language models to discard irrelevant pairs~\citep{Schuhmann2021LAION400MOD}, semantic deduplication to remove redundancy~\citep{abbas2023semdedup}, and prototypicality-based approaches to retain representative data~\citep{Sorscher2022BeyondNS}. 
However, these techniques are less effective for HCS, where redundancy, variability, and subtle morphological differences make conventional filtering challenging. Our work addresses these limitations by building on \citet{Celik2022.12.09.519400}'s \textit{perturbation consistency} framework to curate a balanced dataset of images across semantic classes, which is vital for effective learning under the masked objectives~\citep{zhang2022how}.

\paragraph{Layer-wise Analysis of Deep Neural networks.}
Recent work suggests that intermediate layers (or, blocks) in large ViTs may achieve superior performance on certain linear probing tasks compared to the final encoder layer \citep{evci2022head2toe, dehghani2023scaling}. \citet{alkin2024mim} reported that intermediate layers in large MAE-ViTs (ViT-L, ViT-H) have superior ImageNet-1K \textit{k}\nobreakdash-NN accuracy, likely because later encoder layers become more optimized for the reconstruction task. 

\paragraph{Evaluating Representations for Drug Discovery.} Evaluating the quality of biological representation learning methods for drug discovery remains challenging, as ground truth data is sparse, noisy, biased to well-studied diseases and pathways, and poorly annotated. Metrics have been proposed that use mean average precision \citep{Kalinin_mAP_benchmark} or AUC ROC \citep{Sivanandan_POSH} to assesses how similar related samples are represented, including replicates of the same perturbation or different perturbations with similar annotated biological activities. Recently, \citet{Celik2022.12.09.519400} introduced terminology for describing perturbative “maps of biology”, in which representations of perturbations in HCS data can be placed in unified, relatable embedding spaces allowing for the generation of genome-scale sets of pairwise comparisons. Here we leverage the \textit{biological relationship recall} benchmark proposed by \citet{Celik2022.12.09.519400}, 
which assess how well known relationships between pairs of perturbations are recalled among the most similar or dissimilar embeddings. Computing reliable versions of these relationship benchmarks with HCS data is particularly expensive as they require genome-wide embeddings to be inferred for hundreds of millions of image crops from the genome-wide RxRx3 microscopy screen~\citep{fay2023rxrx3}.

\section{Vision Transformers for Microscopy Images} 
We train and evaluate various vision transformers (ViTs, \autoref{table:models-condensed}) as encoders to extract feature embeddings from $256 \times 256 \times 6$ (HxWxC) microscopy image crops (\autoref{fig:anax_vis}).

\subsection{Training Dataset Curation} \label{sec:data}
Many academic and industry labs have adopted the Cell Painting imaging protocol \citep{cellpainting}, which multiplexes fluorescent dyes to reveal eight broadly relevant cellular components. The datasets used here contain a six-channel implementation of Cell Painting (\autoref{fig:anax_vis}), as well as brightfield images, spanning 100,000s of chemical and genetic perturbations applied to dozens of cell types \citep{kraus2024masked}.
In these datasets, cells that look like unperturbed cells tend to be very over-represented because many perturbations do no induce a morphological change. Some morphological changes are also far more common (e.g. many perturbations will kill cells, resulting in a relatively high proportion of dead cell morphological phenotype). This results in significant imbalance in the morphological phenotypes that the models learn to reconstruct.

To address this, we constructed an 
aggressively curated training dataset (\S~\ref{sec:curation}). 
To learn an initial representation, we began by reproducing the MAE-L/8 model of \citet{kraus2024masked} on a dataset of similar size consisting of 93 million HCS images.  
Using this representation, we first filtered perturbations that did not induce consistent morphological changes to cells. To perform this filtering, we utilized \citet{Celik2022.12.09.519400}'s non-parametric perturbation consistency test (\S~\ref{sec:pert_con}) after correcting for batch effects using Typical Variation Normalization \citep{TVN2017, kraus2024masked}. 
This test was applied within each experiment for computational efficiency, and we restricted the analysis to wells containing single perturbations. 
This consistency was computed for CRISPR guides, siRNAs, and particular concentrations of small molecules across replicates of the same perturbation. P-values were computed for each gene and each (perturbation, concentration) pair. When multiple experiments existed for the same condition, we combined p-values using the Cauchy Combination test \citep{Liu2018CauchyCT}.

We repeated this procedure with a weakly supervised learning (WSL) model trained on RxRx1 \citep{sypetkowski2023rxrx1} and filtered to perturbations where any condition had a p-value $< 0.01$ in either the MAE-L/8 or WSL model. 
This process reduced our original dataset of 93M samples to 16M, which we refer to as Phenoprints-16M. 
While some redundancy remains when distinct perturbations have the same effect, the proportion of samples with that differ from negative controls increased substantially with little decrease in overall diversity. We believe that iteratively repeating this process with  the best models from previous iterations to guide data selection for subsequent models may be a viable strategy. 





\subsection{Models} \label{sec:models}

\paragraph{Baselines.} We compare to several non-finetuned baseline ViT image encoders: three different Dino-v2 backbones \citep{oquab2024dinov2learningrobustvisual} (with 4 register tokens \citep{darcet2024visiontransformersneedregisters}) trained on a curated non-biological natural image dataset; a MAE ViT-L/16 trained on Imagenet-21k~\citep{he2022masked}; and an untrained ViT-S/16. We found that channel-wise self-standardization worked best as the image normalization preprocessing for these baselines, and that the class token was slightly better than the global pool of the patch tokens (except for MAE). Convolutional weights in the patch embedding layer were repeated to embed 6 channel images when using models trained on RGB datasets~\citep{rw2019timm}.

\paragraph{Prior work.} Our primary point of comparison is with respect to the best pretrained foundation model presented by \citet{kraus2024masked}, the MAE-ViT-L/8+ trained on RPI-93M. This \phenomone was trained for approximately 40 epochs, learning from over 3.5 billion image crops, using the L2 mean squared error loss function plus an additional Fourier domain reconstruction loss term.

\paragraph{\phenombeta trained on RxRx3.} We trained a new channel-agnostic MAE \citep{kraus2024masked} ViT-S/16 on the RxRx3 dataset \citep{fay2023rxrx3} for 100 epochs. Channel-agnostic ViTs tokenize each image channel separately with shared patch embedding weights and leverage the dynamic sequence length of transformers with repeated positional encodings to train ViTs that can process images with varying numbers of channels~\citep{ChannelViT,bourriez2024,kraus2024masked}. \citet{kraus2024masked} demonstrate that the large MAEs with 8x8 patch size perform either better or the same as the 16x16 channel-agnostic variants for consistently 6-channel data, so we opted to train standard MAEs for the following two new models since they require fewer tokens at inference time.

\paragraph{\phenomoneretrained trained on \ddaynospace.}  Holding the model backbone constant compared to the MAE-ViT-L/8 by \cite{kraus2024masked}, we assess the impact of our curated dataset in contrast to the 93M dataset by training a new ViT-L/8 MAE for 500 epochs on \ddaynospace. 

\paragraph{\phenomtwo trained on \ddaynospace.} Holding the dataset constant compared to MAE-L/8 above, we assess the impact of increased model scale in terms of parameters by training a new ViT-Gigantic MAE with nearly 1.9 billion parameters for 500 epochs on \ddaynospace. Training this model required 256 H100 GPUs running in parallel for over 1 week. See \S~\ref{sec:appendix_training} for other hyperparameter settings we used for model training.

\begin{figure}[t]
    \centering
    \begin{subfigure}[t]{\linewidth}
        \centering
        \includegraphics[width=\linewidth]{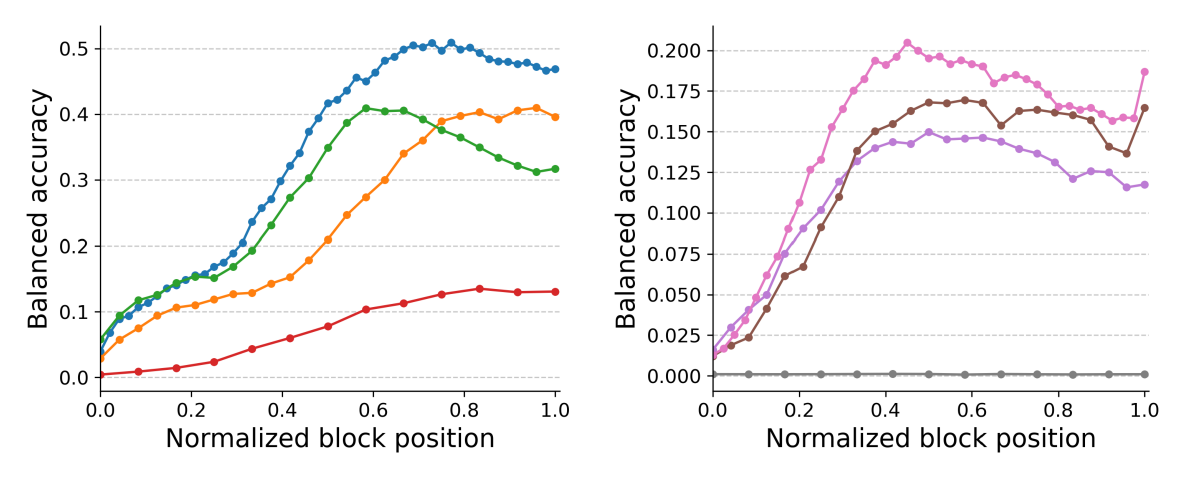}
        \caption{RxRx1 siRNA knockdown classification.}
        \label{fig:rxrx1}
    \end{subfigure}
    
    
    \begin{subfigure}[t]{\linewidth}
        \centering
        \includegraphics[width=\linewidth]{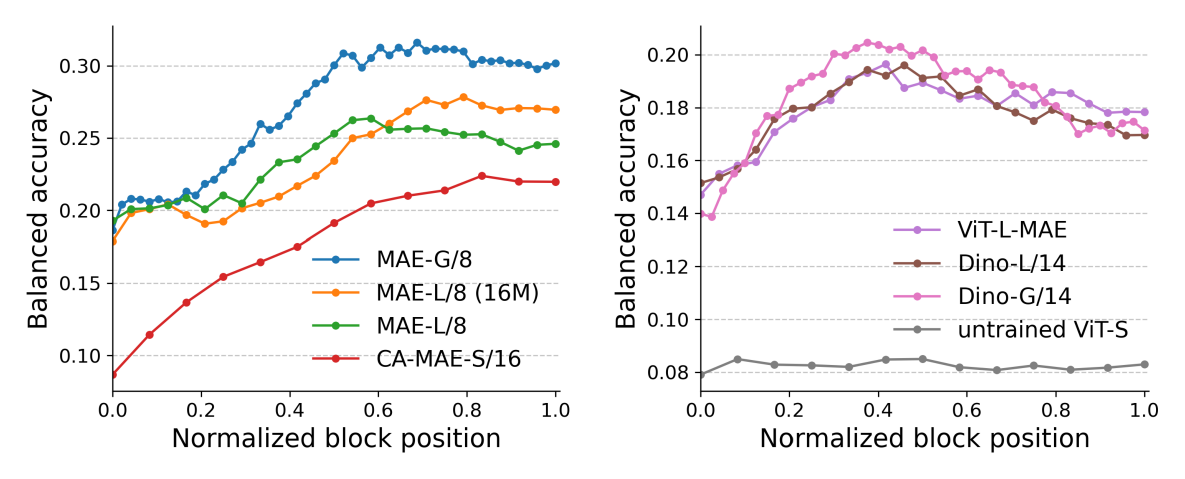}
        \caption{Anax functional gene group classification.}
        \label{fig:anax_huvec}
    \end{subfigure}
    
    \caption{Block-wise validation set \textbf{linear probe results} comparing ViT models pretrained on cell microscopy images (left) versus natural images (right); note the difference in y-axes. (a) 1139-class RxRx1 SiRNA knockdown classification \citep{sypetkowski2023rxrx1}; (b) 40-class Anax functional gene group classification on HUVEC cell images from RxRx3 CRISPR knockouts~\citep{fay2023rxrx3}.}
    \label{fig:consolidated_2x1}
\end{figure}

\section{Linear probing representation learning across ViT blocks} \label{sec:probe_results}

\begin{figure*}[ht]
    \centering
    \includegraphics[width=\linewidth]{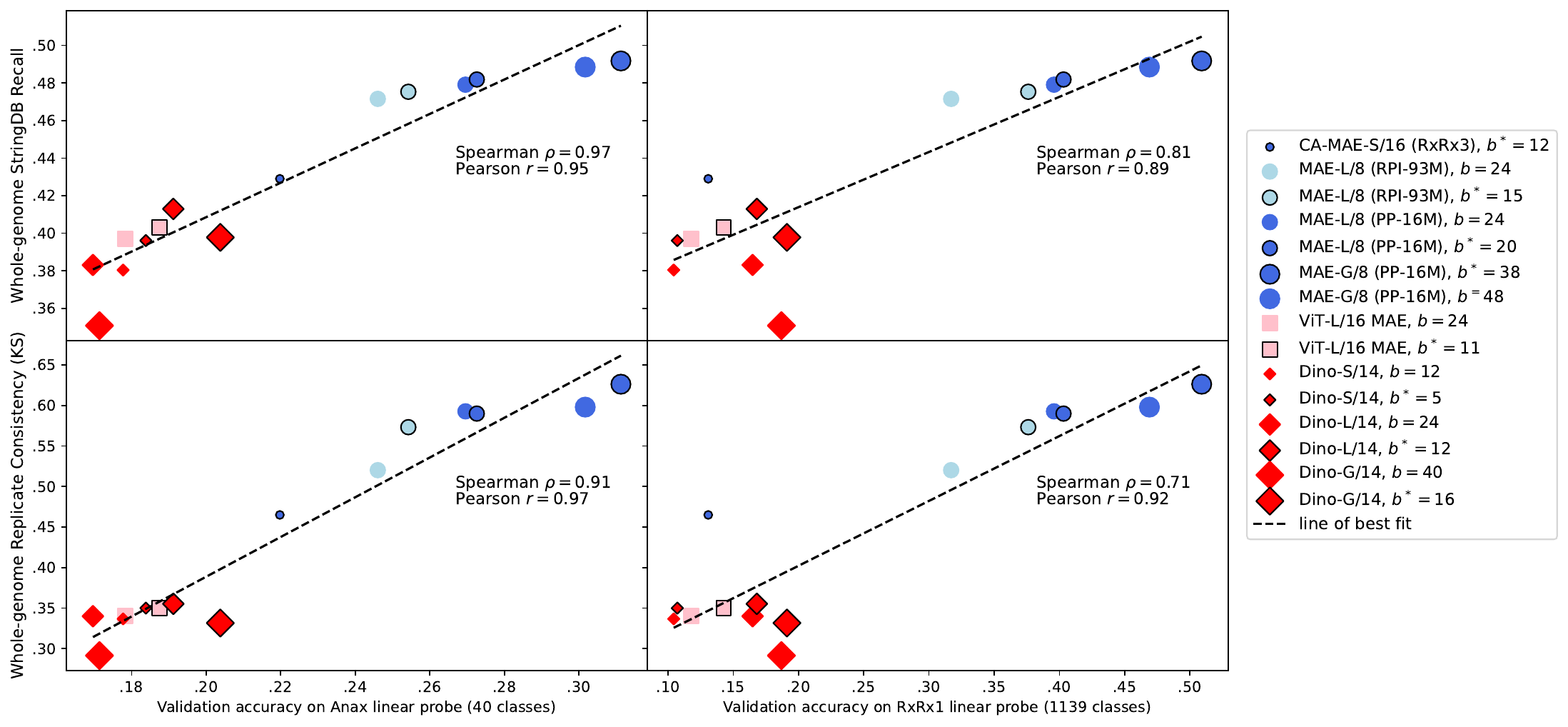}
    \caption{\textbf{Correlations} between validation set linear probing (Figure~\ref{fig:consolidated_2x1})  on Anax and RxRx1 for best and last blocks (Eq.~\ref{eq:optimal_block}) compared to downstream whole-genome benchmarks (Table~\ref{table:bmdbpertcon}) for biological relationship recall on StringDB at 0.05-0.95 gene-gene cosine similarity threshold and replicate consistency KS statistic. Models with \textbf{bold} borders are \textit{trimmed}, \textcolor{red}{red} are natural image baseline models and \textcolor{blue}{blue} are trained on microscopy.
    }
    \label{fig:correlations}
\end{figure*}


We improve the quality of our learned image representations by leveraging previous findings that suggest intermediate blocks within an encoder can provide better representation compared to the final block~\citep{alkin2024mim}. Unfortunately, it is infeasible to search for the best block by simply performing whole-genome evaluation on each block of a large model because the evaluation is extremely time-consuming and resource intensive. For example, evaluating the final block of \phenomtwo required 4,000 L4 GPU hours just for inference (\S~\ref{sec:wholegenome}). We demonstrate that using block-wise linear probes provides insights into the quality of biological features extracted by these models in their intermediate blocks, allowing us to trim the model to an earlier block to both reduce inference costs and improve representation quality.

Our block-wise search consists of training a logistic regression model (linear probe) on the output features of each transformer block to predict either the gene that was perturbed or the functional group that the gene belongs to, and test performance on held-out experiments (\S~\ref{sec:probing}). We define the optimal block $b^*$ for a probing task as the block whose output features achieve the highest test balanced accuracy when trained on the probing task, across all $N$ blocks of the encoder,

\begin{equation}
    b^* = \argmax_{b \in \{1, 2, \dots, N\}} \text{BalancedAccuracy}(\mathbf{z}^{(b)}),
    \label{eq:optimal_block}
\end{equation}

where $\mathbf{z}^{(b)}$ are output features from block $b$ of a ViT. Performance on our linear probing tasks can be viewed as a measure of linear separability of a feature space across experimental batches. 


\paragraph{RxRx1 1139-class siRNA genetic perturbation classification.}
We expect high quality representations of cell images to generate similar embeddings for cells with the same perturbation, hence a simple linear probe should be able to predict gene perturbation from these representation reasonably well. We train linear probes on the publicly-available RxRx1 dataset \cite{sypetkowski2023rxrx1} which consists of 125,510 high-resolution fluorescence microscopy images of human cells under 1,138 siRNA-induced gene knockdowns (plus unperturbed controls) across four cell types (HepG2, HUVEC, U2OS, RPE). These gene knockdowns produce strong phenotypes which makes the prediction task more feasible. 


We found that, for \phenomtwo, the best features came from intermediate block $b^* = 38$ (out of $48$) of the encoder, achieving a balanced accuracy ($0.51$) that is $8.5\%$ greater compared to its final block's output features (\autoref{fig:rxrx1}, left). Additionally, these features achieved $60\%$ greater accuracy than the typically used final block of MAE-L/8+~\citep{kraus2024masked}. We observed similar trends for ViT models pretrained on natural images. For example, DINO-G/14 and ViT-L/16 MAE trained on non-biological natural image data have their best features at blocks that are positioned within the first half of the encoder. For ViT-L/16 MAE, the performance of the best block is $27\%$ higher compared to its final block output features that are typically used for downstream tasks. The higher performance observed for intermediate blocks does not appear to be an intrinsic feature of the ViT architecture as an untrained ViT did not exhibit such a parabolic trend (\autoref{fig:rxrx1}, right).
 

\paragraph{Anax 40-class functional gene group classification.}
Biologically meaningful representation of microscopy images of genetically perturbed cells should capture functional relationships between genes, hence a simple linear probe should be able to predict functional gene groups when trained on these representations. 
We curated a small subset of 80,000 wells from RxRx3 \citep{fay2023rxrx3} to evaluate linear probes on functional group prediction. We also evaluated similar whole genome knockout screens with ARPE-19 and an additional population of HUVEC cells with soluble TNF$\alpha$ added to all wells.
We manually curated Anax, a set of 40 functionally-diverse gene groups containing 348 genes, with details provided  in (\S~\ref{sec:anax_groups}). Examples of groups include major protein complexes (e.g. proteasome, ribosome-small/large), metabolic pathways (e.g. Krebs cycle) and signaling pathways (e.g. calcium signaling) (\autoref{fig:anax_vis}). These groups span broad biological processes that are conserved across cell types -- linear separability of these groups would likely indicate that representations are biologically meaningful regardless of cell type. 

As shown in \autoref{fig:anax_huvec}, \phenomtwo significantly outperforms other models in Anax group linear probe classification. The best representations once again are obtained from an intermediate block, achieving a balanced accuracy ($0.32$) that is $5\%$ greater compared to its final block. We observed similar trends for ViT models pretrained on natural images  and representations computed from microscopy images of other cell types/conditions (\S~\ref{sec:anaxapp}, \autoref{fig:anax_celltypes}).

In \autoref{fig:correlations}, we observe that performance on this novel linear probing task correlates strongly with downstream whole-genome benchmarks across all models (\autoref{table:bmdbpertcon}), whether they are trained on microscopy data or natural images, achieving an overall rank correlation $\rho=0.97$ with whole-genome StringDB recall and $\rho = 0.91$ with whole-genome replicate consistency. This strong correlation is crucial as it allows us to trim our model to the block with the best linear probe performance as a way to improve the quality of our representations for the whole-genome (\autoref{table:bmdbpertcon}).

\section{Whole-genome benchmarking} \label{sec:wholegenome}

\begin{table}[h!]
\caption{Multivariate \textbf{known biological relationship recall} and univariate \textbf{replicate consistency} benchmarks by model, encoding block $b$, aggregated recall over a total of 145,447 possible gene-gene relationships annotations accumulated across five benchmark databases, and replicate consistency test statistics. The \textit{trimmed} models used linear probes to select an earlier block as the feature encoder (Fig.~\ref{fig:consolidated_2x1}). Each benchmark is computed over TVN-aligned gene-aggregated model embeddings, for each one higher is better, and the best overall result is in \textbf{bold}. We report the \textbf{Recall \%} of biological relationships obtained below the 5th and above the 95th percentiles (0.05-0.95) of all  gene-gene cosine similarities against the null distribution, as per \citep{Celik2022.12.09.519400,kraus2024masked}, with mean $\pm$ the standard deviation computed over 3 different random seeds of sampling the null distribution in each benchmark run. For replicate consistency, we report the Komologorov-Smirnov (\textbf{KS}) and Cramer-von Mises (\textbf{CM}) test statistics.}
\centering

\begin{tabular}{lccccc}
\toprule
Model backbone & $b$ & \textbf{Recall} $\%$ & \textbf{KS} & \textbf{CM} \\
\midrule
\multicolumn{1}{l}{\textbf{Baseline ViTs}} \\
ViT-S/16 Untrained &$12$& 34.2  {\scriptsize $\pm$ .08}  & .30 & 4.3 \\
ViT-L/16 ImgNet MAE  &$24$& 37.4  {\scriptsize $\pm$ .06} & .34 & 5.1 \\
\multicolumn{1}{r}{\textit{trimmed}} &$11$& 37.8  {\scriptsize $\pm$ .05} & .35  & 5.8 \\

\midrule
\multicolumn{1}{l}{\textbf{Baseline Dino ViTs}} \\
ViT-S/14 Dino-V2  &$12$& 35.6  {\scriptsize $\pm$ .09} & .34 & 5.6 \\
\multicolumn{1}{r}{\textit{trimmed}} &$5$& 37.2 {\scriptsize  $\pm$ .16} & .35 & 6.0 \\
ViT-L/14 Dino-V2 &$24$& 35.7 {\scriptsize $\pm$ .05} & .34 & 5.3 \\
\multicolumn{1}{r}{\textit{trimmed}} &$12$&  38.8  {\scriptsize $\pm$ .04}  & .36 & 5.9\\

ViT-G/14 Dino-V2  &$40$& 32.7 {\scriptsize  $\pm$ .10} & .29 & 3.8 \\
\multicolumn{1}{r}{\textit{trimmed}} &$16$& 37.5  {\scriptsize $\pm$ .13}
 & .33 & 5.2 \\

\midrule
\multicolumn{1}{l}{\textbf{Microscopy MAEs}} \\
\phenombeta RxRx3 &$12$& 39.6 {\scriptsize $\pm$ .10}  & .47 & 10.4 \\
\phenomone RPI-93M &$24$ & 44.4 {\scriptsize $\pm$ .12}& .52 & 12.3 \\
\multicolumn{1}{r}{\textit{trimmed}} &$15$& 44.3 {\scriptsize $\pm$ .12} & .57 & 15.2 \\
\phenomoneretrained PP-16M &$24$ &44.4 {\scriptsize $\pm$ .12}& .59 & 16.2 \\
\multicolumn{1}{r}{\textit{trimmed}} & $20$ & 44.7 {\scriptsize $\pm$ .06} & .59 & 16.2  \\

\phenomtwo PP-16M &$48$& 45.4 {\scriptsize $\pm$ .07} & .60 & 16.4 \\
\multicolumn{1}{r}{\textit{trimmed}} &$38$& \textbf{45.4} {\scriptsize $\pm$ .15} & \textbf{.63} & \textbf{18.2} \\
\bottomrule
\end{tabular}

\label{table:bmdbpertcon}
\end{table}

\autoref{table:bmdbpertcon} presents our benchmarks computed across the whole-genome. These evaluate the genomic representations obtained for each model by aggregating millions of embeddings of cell images spanning $>$100,000 of genetic knockout perturbations (17,063 genes $\times$ 6 single guide RNAs each) on HUVEC cells from RxRx3 \citep{fay2023rxrx3}. Computing these benchmarks for HCS screens typically requires inferring 140 million crops from the genome-wide RxRx3 microscopy screen \citep{kraus2023masked} (64 tiled crops per each of the 2.2 million wells), but, to reduce compute costs, we discard the outer ring of crops, leaving the 36 center non-edge crops for each well. This requires 80 million forward passes to comprehensively evaluate a new encoder. 
After inference, we use typical variation normalization \citep{TVN2017} and chromosome arm bias correction \citep{lazar2023high} to post-process the embeddings and aggregate them to the gene-level.

We present the multivariate \textbf{biological relationship recall} benchmarks proposed by \citet{Celik2022.12.09.519400} and originally evaluated for MAEs by \citet{kraus2023masked,kraus2024masked}. These metrics evaluate how many annotated pair-wise relationships are recalled from public databases (CORUM, hu.MAP, Reactome-PPI, Signor, StringDB) in the extremities of a ranked list of cosine similarities of all pair-wise post-processed embeddings (details in \S~\ref{sec:A_BMDB}).

To ensure embeddings represent technical replicates of perturbations consistently, we also evaluate model performance on \textbf{replicate consistency} based on the experimental design used in the RxRx3 dataset. Specifically, we compare the similarity of the embedding for corresponding wells across different experiments via a non-parametric statistical test. 
The test statistic measures the difference between the perturbation replicates' similarity distribution and an empirical null distribution, with larger values indicating greater consistency (\S~\ref{sec:rep_con}).
To compare models, we summarize the resulting statistics over all technical replicates in RxRx3 by taking their median, reported in columns KS and CM (Kolmogorov–Smirnov and Cramer-von Mises test statistics) in \autoref{table:bmdbpertcon}. 

Not suprisingly, even the smallest CA-MAE-S/16 trained on microscopy data outperforms the largest baselines ViTs trained on natural images. Training self-supervised on our novel carefully curated \dday  dataset improves the performance of the MAEs, as does trimming to an early layer detected by linear probes, while the most scaled model \phenomtwo achieves the best overall performance in all respects when \textit{trimmed}.
Compared to the best previously SOTA model at its normally used final block, MAE-L/8 RPI-93M \citep{kraus2023masked}, \phenomtwo \textit{trimmed} obtains: (a) the highest statistically significant gain in relationship recall over the previous SOTA, achieving a z-score of improvement of 5.21; (b) a 21\% improvement in the KS statistic (.52$\rightarrow$.63); and, (c) a 48\% improvement in the replicate consistency CM statistic (12.3$\rightarrow$18.2). 

\begin{table*}[t]
    \caption{Comparing MAEs to manually extracted CellProfiler \citep{mcquin2018cellprofiler} features on biological relationship recall. Reported at 0.05-0.95 cosine threshold on the public JUMP-CP image dataset \citep{chandrasekaran2023jump}, which is generated by completely different labs and assay protocols compared to the images used for pretraining. Each result has a standard deviation $\leq \pm .0023$, spanning gene-gene relationships across nearly 8,000 gene-knockouts.}
    \label{tab:jump_bmdb}
    \centering
    \begin{tabular}{lclcccc}
    \toprule
    Model backbone & \textbf{$b$} & Pretraining data & \textbf{CORUM} & \textbf{hu.MAP} & \textbf{Reactome} & \textbf{StringDB} \\
    \midrule
    CellProfiler & - & N/A & .219 & .184 & .131 & .191 \\
    CA-MAE-S/16 & 12 & RxRx3 & .233 & .199 & .154 & .214 \\
    MAE-L/8 & 24 & RPI-93M & .248 & .208 & .160 & .226 \\
    MAE-G/8 \textit{trimmed} & 38 & Phenoprints-16M & \textbf{.264} & \textbf{.215} & \textbf{.165} & \textbf{.235} \\
    \bottomrule
    \end{tabular}

\end{table*}

Linear probing to select optimal ViT blocks leads to significant improvements even when applied to Dino-V2 and MAE models pretrained on natural images. Dino-V2 ViT\mbox{-}G  is dramatically better in terms of recall and consistency with the early block embeddings at $b^*=16$ rather than the final embedding from $b=40$ (which performs worse than a random untrained ViT-S). Dino-V2 ViT-S also observes improvements by using $b^*=5$ rather than $b=12$ and outperforms Dino\mbox{-}V2 ViT-G in replicate consistency, while the \textit{trimmed} Dino-V2 ViT-L obtains the best recall among the baselines. This finding is important for the scientific community to consider when applying ML foundation models to experimental data, as it is common to take embeddings from the final layer as a default strategy even when processing potentially out-of-distribution images \cite{lastufka2024vision}, which, as we have shown, can significantly hinder results.


\subsection{Performance on external data (JUMP-CP)} \label{sec:jump}
In order to validate that the MAEs generalize to entirely novel data, we evaluated a subset of models on completely external public data generated by different assays and from a variety of different labs as produced by the JUMP-CP consortium \citep{chandrasekaran2023jump}. \autoref{tab:jump_bmdb} presents these results using the relationship recall benchmarks of \citep{Celik2022.12.09.519400}, noting that only a subset of 7,976 gene-knockouts are covered by this dataset. For post-processing embeddings, we use PCA with center-scaling for standard batch correction alignment. We observe that the MAEs perform better than the CellProfiler manual feature extraction baseline (\S~\ref{sec:cellprofiler}) \citep{carpenter2006cellprofiler,kamentsky2011improved}, and that the general trend is maintained with the trimmed MAE-G/8 obtaining the best recall overall. Notably, recall on JUMP-CP is considerably lower than on RxRx3 (\autoref{table:bmdbpertcon}) likely due to different assay protocols and more variance in the data.

\section{RxRx3-core benchmarking} \label{sec:rxrx3-core}

RxRx3-core\footnote{\url{https://huggingface.co/datasets/recursionpharma/rxrx3-core}} \citep{kraus2025rxrx3} is a publicly available benchmarking dataset for assessing biological capabilities of computer vision models. RxRx3-core includes labeled images (compressed to JPEG-2000) of 735 genetic knockouts and 1,674 small-molecule perturbations across eight concentrations drawn from 222,601 wells ($512 \times 512 \times 6$ pixel center-crops)  drawn from the larger RxRx3 dataset. 

We evaluate a random choice baseline, CellProfiler, the CA\mbox{-}MAE\mbox{-}S/16 model, the MAE-L/8 model from previous work \citep{kraus2023masked}, and the MAE-G/8. We evaluate both the trimmed and full-length version of the latter two models to determine the impact of our model-trimming strategy for the most performant models in this context. 

We also evaluate a preliminary version of a SSL ViT-L/16 we trained with the Dino-V2 algorithm on the RxRx3 microscopy dataset; however, we found that it underperformed compared to the CellProfiler 4 baseline (\S~\ref{sec:cellprofiler}) \citep{kamentsky2011improved}  and, unlike the MAEs we trained, significantly overfit during pretraining (\S~\ref{sec:dinov2}). As such, we did not pursue additional analysis on this model. We leave further development of Dino-V2 for microscopy data to future work, as we were unable to determine an effective recipe for applying Dino SSL pretraining on large-scale microscopy data.

To evaluate each model, we first inference all 222,601 $\times$ 4 crops and then average the 4 embeddings to the well-level. Then, to perform standard batch correction alignment, we use the ``EMPTY'', unperturbed wells as our control population. We fit PCA on those control embeddings, use it transform the rest, and then fit a separate standard-scaler on each batch's controls to transform the rest. This simplified alignment strategy empirically performed better than TVN on this dataset.

\begin{figure*}[t]
    \centering
    \includegraphics[width=\linewidth]{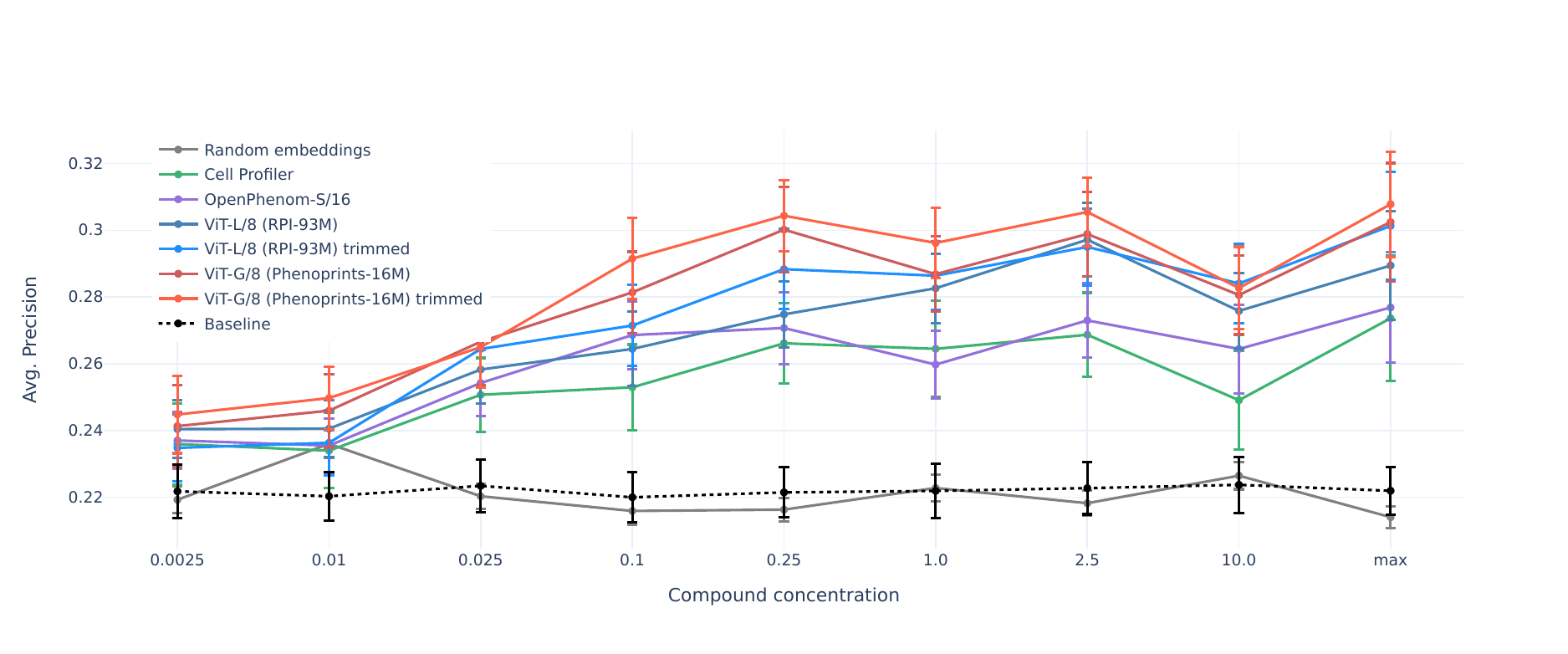}
    \caption{Mean average precision performance on RxRx3-core public benchmark in predicting compound activity against annotated gene targets, across all compound concentrations with error bars for 100 runs of the benchmark with different random seeds (\autoref{tab:compound_gene_rxrx3core}).}
    \label{fig:compounds}
\end{figure*}

We present results for the benchmark measuring zero-shot prediction of compound-gene activity using cosine similarities between embeddings (\autoref{fig:compounds}). This measures, for each compound, whether the cosine similarities from a model's embeddings correctly rank the compound's known target genes higher than a randomly sampled set of other genes from a ground truth dataset. \autoref{tab:compound_gene_rxrx3core} provides exact values along the \textit{max} axis, which captures the strongest potential interaction regardless of concentration. The relative ranking of model performance, holds as expected from the results in \S~\ref{sec:probe_results} and \S~\ref{sec:wholegenome}, and trimming benefits both MAE-L/8 and MAE-G/8, with MAE-G/8 offering a 42\% (3.77 $\rightarrow$ 5.38) relative improvement over the method from previous work in predicting compound-gene activity.

\begin{table}[t]
    \caption{Performance on public \textbf{RxRx3-core compound-gene benchmark}, measuring mean ($\pm$ STD over 100 random seeds) average precision in predicting compound activity against target genes with its z-score of improvement over the random baseline.}
    \label{tab:compound_gene_rxrx3core}
    \centering
    \begin{tabular}{lcc}
        \toprule
        \textbf{Model} & \textbf{Avg. prec.} $\uparrow$ & \textbf{Z-score} $\uparrow$ \\
        \midrule
       \textit{Random baseline} & 0.222 {\scriptsize $\pm$ .007}  & 0.00 \\
       CellProfiler & 0.274 {\scriptsize $\pm$ .019} & 2.55\\
       
       Dino-V2 ViT-L/16, RxRx3 & 0.258 {\scriptsize $\pm$ .015} & 2.13\\    
       CA-MAE-S/16, RxRx3 & 0.273 {\scriptsize $\pm$ .016}  & 2.90 \\
       MAE-L/8, RPI-93M & 0.290 {\scriptsize $\pm$ .017} & 3.77 \\
       \multicolumn{1}{r}{\textit{trimmed}} & 0.299 {\scriptsize $\pm$ .016} & 4.49 \\
       MAE-G/8, PP-16M & 0.302 {\scriptsize $\pm$ .015}  & 4.79 \\
       \multicolumn{1}{r}{\textit{trimmed}} & \textbf{0.309} {\scriptsize $\pm$ .015}  & \textbf{5.38} \\
        \bottomrule
    \end{tabular}

\end{table}

\section{Discussion and Conclusions}

This results in this work demonstrate that: 
(1) within the context of biological imaging, trimming many ViTs to an earlier block leads to stronger biological linearity and improved performance on downstream tasks in addition to cheaper inference costs (\autoref{fig:consolidated_2x1}); (2) linear probing performance on a subset of genetic perturbations correlates strongly with downstream performance on whole-genome benchmarks and can be used to optimize which block is selected for representing the whole-genome (\autoref{fig:correlations}); (3) the most scaled model, \phenomtwo trained on the specially curated dataset Phenoprints-16M for 500 epochs, obtains the overall best performance across all benchmarks and linear probes, providing further evidence for the scaling hypothesis in biological image data (\autoref{table:bmdbpertcon}, \autoref{tab:jump_bmdb}, \autoref{tab:compound_gene_rxrx3core}, \S~\ref{sec:FLOPS}). 
We have found that intentionally scaling training compute and parameters of MAEs for microscopy on curated data can benefit a wide variety of biologically relevant tasks, even in comparison to a model trained on a $5\times$ larger dataset. Indeed, even after training the MAE-G/8 on more than eight billion crops from a 16-million-image dataset, the validation reconstruction loss continued to improve. This suggests that increasing the model’s parameters further on this data may continue to yield improvements.


More broadly, this work proposes a reusable recipe for training and extracting optimal representations from fully self-supervised models trained on experimental data. The pattern we use can be applied to other domains that contain data from repeated experiments but without accurate ground truth labels. Specifically, we recommend: (a) curating the training set by identifying diverse sets of samples that are represented consistently, e.g., by using a pre-existing model to select such samples; (b) training a scaled transformer-based model using a self-supervised learning technique, such as masked autoencoding; and, (c) inspecting the performance of the trained transformer, and all baselines, at every block to identify the optimal layer for representing the data.    

\section*{Limitations and reproducibility}
In this work, we evaluated publicly available SSL ViTs as baselines and trained new MAE models trained on a specially curated microscopy dataset. Our preliminary attempts to train ViTs with Dino on this microscopy data encountered suboptimal performance (\S~\ref{sec:rxrx3-core}). Consequently, we allocated our time and compute budget to investigate scaling MAE to ViT-G/8 on curated data.
%
We recognize the potential for other SSL pretraining regimes and fine-tuning strategies  \citep{singh2023effectiveness,lehner2024contrastive,hondru2024masked,khan2024best,alkin2024mim} oriented for microscopy data to lead to future improvements on these tasks. We publicly release the inference, reconstruction visualization and benchmarking code\footnote{\url{https://github.com/recursionpharma/maes_microscopy}}, along with the full weights for CA-MAE ViT-S/16\footnote{\url{https://huggingface.co/recursionpharma/OpenPhenom}}.

\section*{Impact Statement}
This paper presents work whose goal is to advance the field of Machine Learning in its applications to the sciences. There are many potential societal consequences of our work, especially relating to the discovery of new biological relationships between genes, potential drug treatments for diseases, and overall accelerating the process of drug development. At the same time, the predictions of these models are not guaranteed to be correct, which is why utmost care must be taken to validate safety and efficacy in orthogonal assays and early-stage clinical trials.



\bibliography{iclr2024_conference}
\bibliographystyle{icml2025}

\appendix
\clearpage
\section{Appendix}  \label{sec:appendix}


\subsection{Training dataset curation details} \label{sec:curation}

In order to produce Phenoprint-16M, we curated 93M using the following steps:
\begin{enumerate}
    \item Filtering out data that did not pass data quality filters related to the focus of the image, quantity of dead cells, assay conditions, and presence of strong anomalous imaging artifacts. 
    \item Filtering out data with missing information about the perturbations applied, data with more than 3 perturbations applied, and data of unusual size (in the image dimension or number of channels).
    \item Filtering out perturbation conditions that had been in less than 3 distinct experiments or 20 distinct wells so as to capture a variety of batch effects and have a broad sample of positives per class.
    \item Under-sampling perturbation conditions that were clearly over-represented in the dataset. Our experiment designs contain positive controls, negative controls, and wells without perturbation within each experiment. At this step, we keep 10\% of positive controls and wells without any perturbation, 30\% of negative controls, and all other perturbation conditions.
    \item Filtering out wells where none of the perturbation conditions had a phenoprint (\S\ref{sec:pert_con}) (across different map types) in any experiment it had been run in. 
\end{enumerate}

\subsection{Training hyperparameters} \label{sec:appendix_training}

\begin{table*}[t]
\centering
\caption{Overview of vision transformer (ViT) encoders used and evaluated in this work.} 
\begin{tabular}{lcccc}
\toprule
\textbf{Model Name} & \textbf{Parameters} & \textbf{Blocks} & \textbf{Model Dim} & \textbf{Pretraining Data}  \\
\midrule
\multicolumn{5}{l}{\textbf{Baselines}} \\
Untrained ViT-S/16 & 25M & 12 & 384 & N/A  \\
Dino-V2 ViT-S/14 & 25M & 12 & 384 & Natural images \\
Dino-V2 ViT-L/14 & 307M & 24 & 1024 & Natural images \\
Dino-V2 ViT-G/14 & 1,100M & 40 & 1536 & Natural images \\
ViT-L/16 MAE & 307M & 24 & 1024 & Imagenet-21k \\
\midrule
\multicolumn{5}{l}{\textbf{MAEs for microscopy}} \\
\phenombeta & 25M & 12 & 384 & RxRx3 \\
\phenomone & 307M & 24 & 1024 & RPI-93M \\
\phenomoneretrained & 307M & 24 & 1024 & \ddaynospace  \\
\phenomtwo & 1,860M & 48 & 1664 & \ddaynospace  \\
\bottomrule
\end{tabular}
\label{table:models-condensed}
\end{table*}

\begin{table*}[ht]
    \centering
    \caption{Training hyperparameters for the new models presented in this work. Each used a one-cycle cosine learning rate decay schedule with 10\% warm-up using the Lion optimizer from \cite{chen2023symbolic} with betas (0.9, 0.95) and weight decay of 0.05, with additional ViT settings such as LayerScale as proposed by \cite{dehghani2023scaling}. $^*$Note that \phenomtwo had multiple restarts during training due to challenges associated with massive model training on large-scale shared distributed compute clusters.}
    \begin{tabular}{lccc}
    \toprule
    \textbf{Hyperparameter} & \phenombeta & \phenomoneretrained & \phenomtwo \\
    \midrule
    Vision transformer backbone & ViT-S & ViT-L & ViT-G \citep{zhai2022scaling}\\
    Pretraining Data & RxRx3 & \ddaynospace & \ddaynospace \\
    Training epochs   & 100   & 500    & 500$^*$ \\
    Learning rate     & 1e-4  & 3e-5   & 3e-5  \\
    Global batch size & 2048  & 16384  & 8192 \\
    Stochastic depth  & 0.1   & 0.3    & 0.6   \\
    \# GPUs           & 16 A100s & 128 H100s & 256 H100s\\
    \# GPU-hours      & 400   & 15,360  & 48,000 \\
    \bottomrule
    \end{tabular}
    \label{tab:model_cards}
\end{table*}

\autoref{tab:model_cards} provides the hyperparameters used for training the new vision transformers presented in this work. Each model was trained using a 75\% mask ratio and the standard decoder architecture for MAEs \citep{he2022masked}. Each model was trained with the standard L2 MAE loss and the Fourier-space loss function implemented by \cite{kraus2024masked} with a weight of $\alpha=0.01$. We note, however, that the details presented by \cite{kraus2024masked} do not precisely correspond with the implementation provided in their Github repository; when reshaping the tokens to a shape compatible with the 2D Fourier transform, the permute operation resulted in adjacent pixels being from different channels of the input, resulting in the high frequency components of the loss being a function of the relationships between input channels. An initial investigation with a ViT-L/8 showed that changing the implementation to the one described in the paper did not dramatically change probing results. As such, we used the implementation as-is and leave additional analysis of loss function design for MAEs to future work.

\subsection{Perturbation Consistency} \label{sec:pert_con}

In order to assess the consistency of the induced morphology on the cells by the perturbations, we used a non-parametric perturbation consistency test similar to the one introduced in \cite{Celik2022.12.09.519400}. Let $x_{g,1}, x_{g,2},\cdots, x_{g,n}$ be the embeddings for replicates of perturbation $x_g$ on experiment (batch) $e$. As the test statistic for perturbation consistency, $\bar{s}_g^{e}$ is defined as the mean of the cosine similarities across all pairs of replicates of $x_g$.

\begin{equation}
    \bar{s}_g^{e} = \frac{1}{n^2} \sum_{i = 1}^ n \sum_{j=1}^n \frac{\langle x_{g,i},x_{g,j}\rangle}{|| x_{g,i}|| ||x_{g,j}||}.
\end{equation}

where $\langle . \rangle$ and $||.||$ denote dot product and $L_2$ norm. 

Statistical significance of $\bar{s}_g^{e}$ is assessed using a permutation test comparing it against an empirical null distribution generated using the same statistic for a set of randomly selected perturbations in experiment $e$, $\{\bar{s}'_1, \cdots,\bar{s}'_K\}$. The p-value for $\bar{s}_g^{e}$ is computed as follows

\begin{equation}
    p_g = \frac{\max \biggl\{ \# \{\bar{s}'_k \geq \bar{s}_g^{e}  \} ,1\biggl\}}{K}.
\end{equation}

When multiple experiments existed for the same perturbation, we combined p-values using the Cauchy Combination test \citep{Liu2018CauchyCT}.

\subsection{Training linear probes} \label{sec:probing}

In this section, we provide details about the training process and preprocessing steps used in our logistic regression models. These models were trained on output features derived from various Vision Transformer (ViT) blocks.

The data was split by experiments, ensuring that the test data originated from experiments distinct from those used for training. This approach helps to validate the generalization performance of our models across different experimental conditions.

For both RxRx1 gene prediction and Anax group prediction, we apply \texttt{StandardScaler} from the scikit-learn library as the only preprocessing step to standardize the features prior to training linear probes. \texttt{StandardScaler} transformation was fitted on data from the train split.
We trained the logistic regression models using scikit-learn’s \texttt{LogisticRegression} class. The following parameters and settings were used during model optimization:
\begin{itemize}
\setlength\itemsep{0.1em}
    \item Solver: lbfgs
    \item Maximum Iterations: 2000
    \item Class Weight: balanced
\end{itemize}
For RxRx1 gene prediction, we trained logistic regression models to predict one of $1139$ possible perturbation labels ($1138$ genetic perturbation and non-perturbed control). For Anax group prediction, we trained logistic regression models to predict one of $40$ possible function group labels (\S~\ref{sec:anax_groups}). We report the balanced test accuracy as the main evaluation metric for all linear probing experiments. 

\subsection{Anax classification for other cell lines/treatment conditions: ARPE19 and HUVEC with TNF$\alpha$ background} \label{sec:anaxapp}

We performed linear probing on imaging data obtained for a retinal pigment epithelia (RPE) cell line, ARPE19, and HUVEC cells treated with an inflammatory cytokine, TNF$\alpha$. We similarly observed that intermediate blocks often have the most linearly separate features compared to the final block. 

\begin{figure}[ht]
    \centering
    \includegraphics[width=0.49\linewidth]{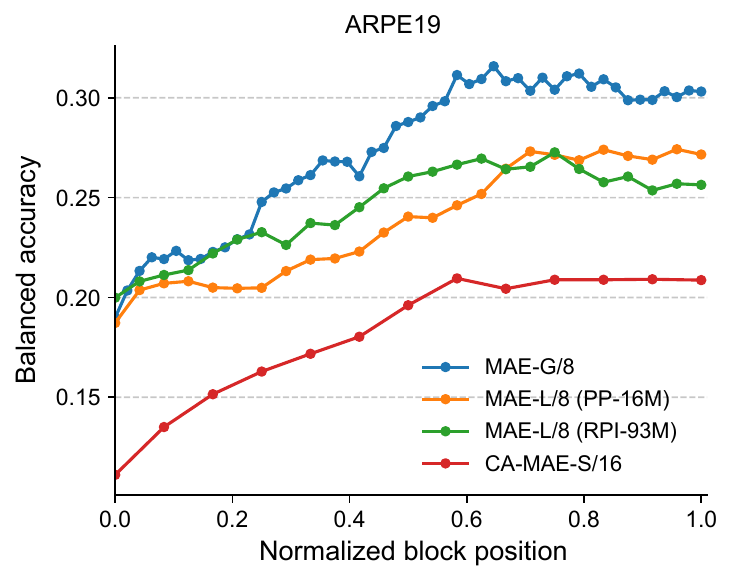}
    \includegraphics[width=0.49\linewidth]{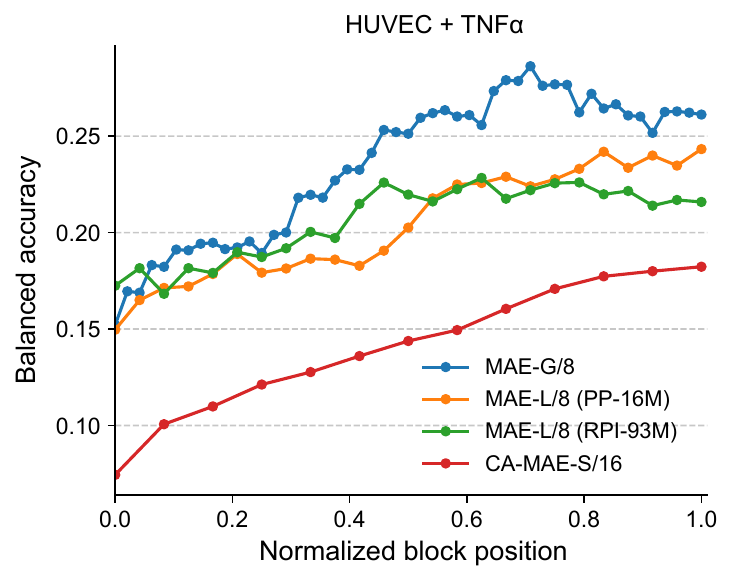}
    \caption{Layerwise validation set linear probe performance on Anax functional gene group classification beyond RxRx3: CRISPR knockouts in the ARPE-19 immortalized epithelial cell-line (left), and in HUVEC cells with a TNF$\alpha$ background (right).}
    \label{fig:anax_celltypes}
\end{figure}

\subsection{Biological Relationship Recall} \label{sec:A_BMDB}
A valuable use of large-scale HCS experiments is to perform large-scale inference of biological relationships between genetic perturbations.
We evaluate each model's ability to recall known relationships by using the \textit{biological relationship recall} benchmark described in \citet{Celik2022.12.09.519400}. First, we correct for batch effects using \textit{Typical Variation Normalization} (TVN) \citep{TVN2017}, and also correct for possible chromosome arm biases known to exist in CRISPR-Cas9 HCS data \citep{lazar2023high}. To infer biological relationships, we compute the aggregate embedding of each perturbation by taking the spherical mean over its replicate embeddings across experiments.
We use the cosine similarity of a pair of perturbation representations as the relationship metric, setting the origin of the space to the mean of negative controls.
We compare these similarities with the relationships found in the following public databases: CORUM~\citep{CORUM}, hu.MAP~\citep{hu_map}, Reactome~\citep{REACTOME}, and StringDB~\cite{stringdb} (with $>$95\% combined score). Table~\ref{table:bmdbpertcon} reports the recall of known relationships amongst the top and bottom 5\% of all cosine similarities between CRISPR knockout representations in RxRx3 \citep{fay2023rxrx3}.

\subsection{Replicate Consistency} \label{sec:rep_con}

In order to assess the reproducibility of the perturbations across their technical replicates, we compare the distributions of the similarities for same perturbations across replicates against an empirical null distribution. 
Specifically, for technical replicate experiments $e^i_a$ and $e^i_b$, we calculate the cosine similarity between the embeddings of perturbation $x_j$ in them, denoted as $s^{x_{j}}$.The query distribution $q^{e_i}$ is constructed by computing the cosine similarities for all perturbations that have a matching well on experiments $e^i_a$ and $e^i_b$. An empirical null distribution of identical cardinality is created by computing cosine similarity, $r^{x_{k},x_{l}}$, between random pairs from $e^i_a$ and $e^i_b$ such that no pair corresponds to the same perturbation, $p^{e_i}_0$. Using non-parametric statistical tests, namely Kolmogorov-Smirnov (KS) and  Cramer Von-Mises (CVM), we can evaluate the hypothesis that $q^{e_i}$ and $p^{e_i}_0$ are drawn from the same distribution. Formally, let  $Q^{e_i}(x)$ and $P^{e_i}_0(x)$ be the cumulative distribution functions for $q^{e_i}$ and $p^{e_i}_0$ respectively, then the KS statistic for the two-sample case of technical replicate experiments $e^i_a$ and $e^i_b$ is defined as:

\begin{equation}
\text{KS}^{e_i} = \text{sup}_{x}|Q^{e_i}(x) - P^{e_i}_0(x)|.
\end{equation}
The Cramér–von Mises test statistic (CVM) for experiments $e^i_a$ and $e^i_b$ is computed as:
\begin{equation}
\text{CVM}^{e_i} = \frac{1}{2N^2}\sum_{m=1}^N\biggl[(r_m - m)^2 + (s_m -m)^2\biggl] - \frac{4N^2 - 1}{12N}.
\end{equation}
where $N$ is the cardinality of $q^{e_i}$ and $p^{e_i}_0$ and $s_m$ and $r_m$ are ranks of similarities $s^{x_{j}}$ and $r^{x_{k},x_{l}}$ in the combined distribution of $q^{e_i}$ and $p^{e_i}_0$ when ordered. In order compare models, we use the median of $\text{CVM}^{e_i}$ and $\text{KS}^{e_i}$ over all technical replicate experiment pairs $e_i$. 

Since the pairs are randomly selected for $p^{e_i}_0$, the embeddings would be mostly orthogonal thus the distribution would be centered around 0.
 Given that not all CRISPR knockouts would induce a morphological change in the cells, it's plausible for distribution $q^{e_i}$ to exhibit a peak around 0. As the model approaches the precision of an oracle, we would anticipate the mass situated around this peak to shift towards higher cosine similarity values.


\subsection{Anax Group Prediction Details} \label{sec:anax_groups}

The Anax probing task introduced in this paper is intended to balance capturing a diverse range of biology that is broadly conserved between cell types with a reduced cost of execution. The name ``Anax'' is a reference to Anaximander, the 6th century B.C. philosopher credited with making the first world map. 

In curating these genes, we analyzed the sources listed in \S~\ref{sec:A_BMDB} as well as internal gene expression data to produce ``functional'' groups corresponding to biological processes, cellular components, and molecular functions. Not all genes within each group are expected to have the same knockout phenotype, but are classified by humans as having related function -- linear separability of these genes would indicate that a model has learned similar concepts to those deemed significant by biologists.

The gene groups we use for the 40-class Anax group classification task (\S~\ref{sec:probing}) are listed in \autoref{tab:anax_group_genes}.

\begin{figure}
    \centering
    \includegraphics[width=\columnwidth]{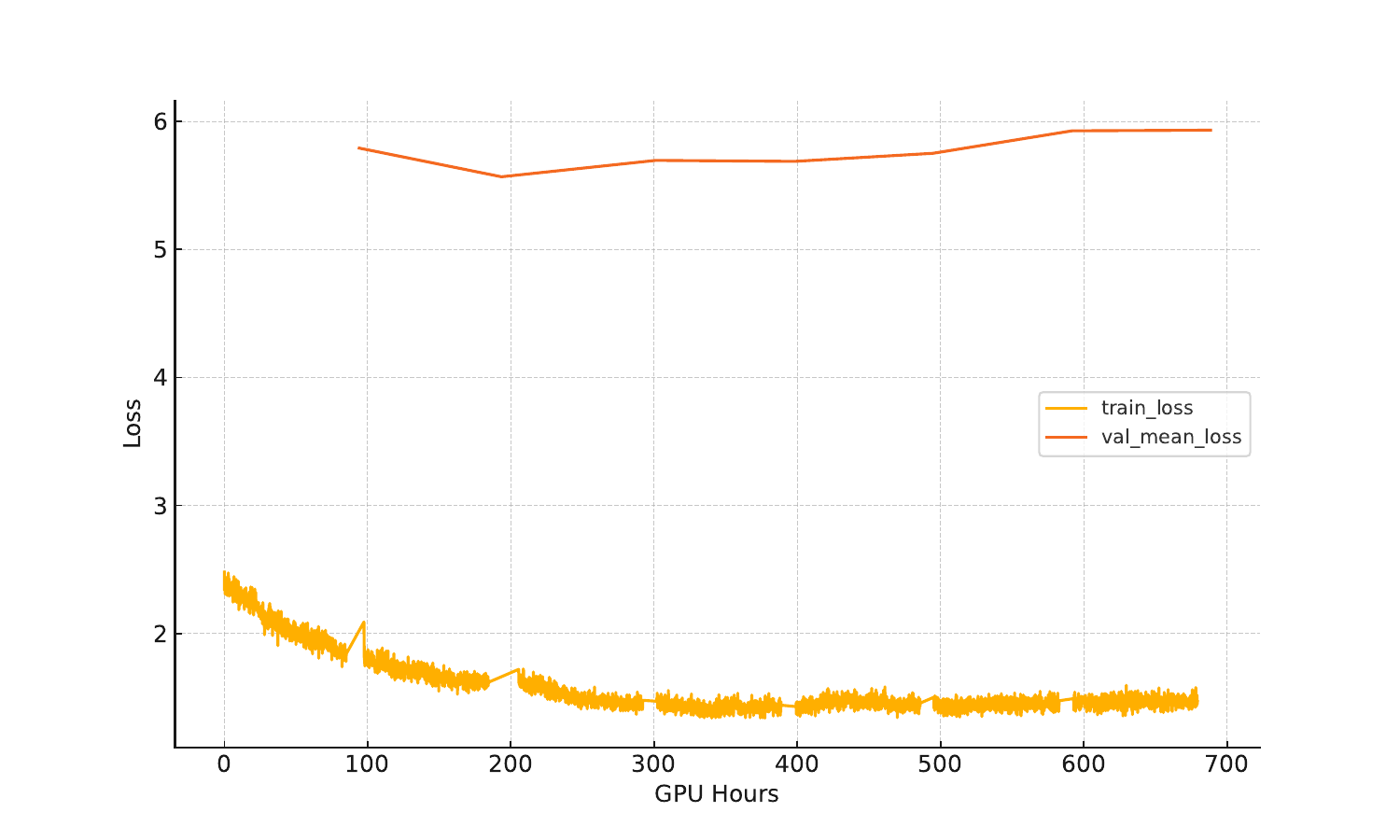}
    \caption{Loss curve when training Dino-v2 ViT-L/16 on RxRx3.}
    \label{fig:dinov2loss}
\end{figure}

\subsection{Dino-V2 pretraining on microscopy data}
\label{sec:dinov2}

We attempted to train two Dino-v2 models on microscopy data. One ViT-L/16 from scratch on RxRx3, and another attempt of fine-tuning the MAE-L/8 on RPI-93M with the Dino-v2 losses. They had the following hyperparameter settings which were tuned on another dataset: output-dim 65536, 2 global crops, 4 local crops, dino loss weight 1.0, koleo loss weight 0.1, ibot loss weight 1.0, Lion optimizer with max learning rate 1e-5, weight decay 0.05, betas 0.9 0.95, and cosine annealing. In both cases, we observed significant over-fitting of the loss from the start (\autoref{fig:dinov2loss}).

    

\begin{table}
    \caption{RxRx3-core benchmarks for our initial attempts to train Dino-V2 models on microscopy data. The latter was finetuned from the MAE-L/8 trained on RPI-93M. Results compare to \autoref{tab:compound_gene_rxrx3core}.}
    \centering
    \begin{tabular}{lcc}
    \toprule
    \textbf{DinoV2 model} &\textbf{ Avg. Prec.} & \textbf{Z-score}\\
    \midrule
        ViT-L/16 RxRx3 
            & 0.258 $\pm$ .015 & 2.13                             
        \\ViT-L/8 (ft.) RPI-93M 
            & 0.255 $\pm$ .018 & 1.76\\
    \bottomrule
    
    \end{tabular}

    \label{tab:dino-microscopy}
\end{table}

In \autoref{tab:dino-microscopy} we show that both models fail to improve on the RxRx3-core benchmark metrics (i.e., z-scores over the random baseline) versus the CA-MAE-S/16 RxRx3 model which had Z-score of 2.90 on average precision for predicting compound activity.
We have not found an effective recipe for training Dino on microscopy data. 
However, we note that theoretical evidence exists arguing that MAE learning is in some ways equivalent to contrastive learning \cite{hondru2024masked}, so even if an appropriate Dino recipe is found it would remain to be seen if it differs substantially from MAEs for microscopy given the same training compute. As described in the \textbf{Limitations} section, we expect that future work would have to dedicate significant training ablations and creativity to determine the best possible training recipe for training Dino on microscopy data.

\subsection{Correlation between model scale and benchmark results}
\label{sec:FLOPS}

\begin{figure*}
    \centering
    \includegraphics[width=\linewidth]{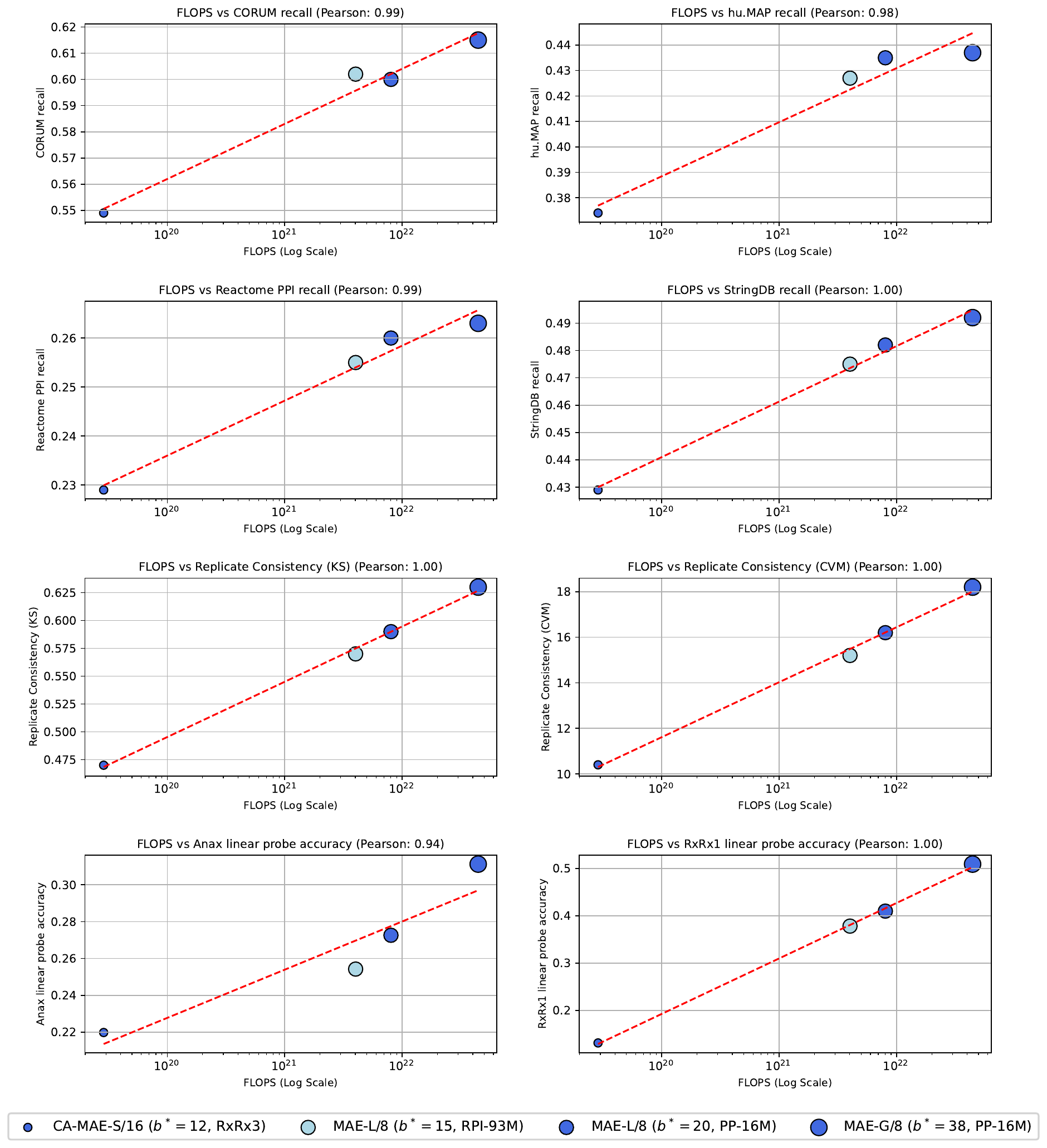}
    \caption{Relationship between FLOPs and benchmark evaluation results for the six whole-genome tasks (\autoref{table:bmdbpertcon}) and the two linear probing tasks (\autoref{fig:consolidated_2x1}).}
    \label{fig:FLOPs}
\end{figure*}

In \autoref{fig:FLOPs} we show the correlations between training FLOps (floating point operations) and downstream results. Over all benchmarks we observe a very strong consistent linear trend where scaling training FLOps improves overall pwerformance. This work provides the next log step in scale as we enter into the billion-parameter model regime with MAE-G/8. These results therefore provide additional evidence that the trend initially discovered by \cite{kraus2023masked} between FLOps and relationship recall actually extends both to billion-parameter models and even moreso for other biologically meaningful benchmarks pertaining to linear probes on small experiments and to replicate consistency on the whole-genome.



\subsection{CellProfiler features}
\label{sec:cellprofiler}

CellProfiler bioimage analysis software \cite{carpenter2006cellprofiler,mcquin2018cellprofiler} was used to compute features using classical segmentation and feature extraction algorithms. Benchmarking results using CellProfiler features are reported for JUMP-CP (\S~\ref{sec:jump}) and RxRx3-core (\S~\ref{sec:rxrx3-core}). JUMP-CP CellProfiler features were downloaded from \href{https://cellpainting-gallery.s3.amazonaws.com/index.html#cpg0016-jump}{https://cellpainting-gallery.s3.amazonaws.com/index.html\#cpg0016-jump}. CellProfiler features for RxRx3-core were computed using  version 2.2.0 \citep{kamentsky2011improved}. Single cells were segmented after applying illumination correction and shape, intensity, and texture features were extracted from each cell, resulting in 952 dimensional profiles. For RxRx3-core, we kept only cells in the center 512x512 crop of original 2048x2048 images and then mean aggregated them. 4,749 wells of the 222,601 RxRx3-core wells were missing CellProfiler features, so these were held out of the benchmark computations for all models on RxRx3-core.

\begin{table*}
\caption{Anax groups and their associated genes. This table presents a comprehensive list of gene groups and their corresponding genes.}
\tiny
\SetTblrInner{rowsep=1pt} 
\begin{tblr}{
  width = \textwidth,
  colspec = {X[0.25,l] X[0.7,l]}, 
  row{1} = {font=\bfseries}, 
  hlines, vlines 
}
Anax Group  & Genes \\
Acyl Coa Biosynthesis & ELOVL2, ELOVL5, ELOVL6, HACD1, HACD2, HSD17B12, SCD, SCD5, TECR \\
Adherens Junctions & ACTB, ACTG1, AFDN, CDH1, CTNNA1, CTNNB1, CTNND1, NECTIN1, NECTIN3, NECTIN4 \\
Amino Acid Metabolism & ALDH4A1, ARG2, CKB, CKMT2, CPS1, DAO, OTC, PYCR2, PYCR3, SAT1 \\
Apoptosis & CFLAR, DFFB, CASP6, CASP3, FASLG, BCL2, DFFA, XIAP, TNFSF10, AKT3 \\
Autophagy & ATG12, ATG3, ATG4B, ATG4C, ATG7, GABARAP, PIK3C3, PIK3R4, PRKAA1, ULK1 \\
Beta Oxidation Of Fatty Acids & ACAA2, ACADL, ACADM, ACADS, ACADVL, ECHS1, ECI1, HADH, HADHA, HADHB \\
Calcium Signaling & ADCY1, ADCY2, ADCY3, CALM1, CAMK2B, CAMK2D, PDE1B, PDE1C, PRKACG, PRKX \\
Clathrin Coated Vesicles & AP2A1, AP2A2, AP2B1, AP2M1, AP2S1 \\
COPI & ARCN1, COPA, COPB1, COPB2, COPE, COPG1, COPZ1 \\
COPII Vesicles & SEC13, SEC23A, SEC24B, SEC24D, SEC31A \\
DNA Damage Repair & BLM, BRCA2, EME1, NBN, POLD2, RAD51B, RAD51C, RAD51D, RPA1, XRCC2 \\
Dynein & DYNC1H1, DYNC1I2, DYNC1LI1, DYNC1LI2, DYNLT1 \\
ER Protein Translocation & SPCS3, SEC61A1, SRP14, SRP72, SPCS1, SRPRA, SEC11A, SRP68, SRPRB, SRP54 \\
Exosome & DIS3, EXOSC10, EXOSC3, EXOSC4, EXOSC5, EXOSC6, EXOSC7, EXOSC8, EXOSC9, MPHOSPH6 \\
Gap Junctions & ADCY8, DRD2, HTR2C, ITPR2, LPAR1, PDGFD, PDGFRB, PLCB3, TUBA1C, TUBB1 \\
Golgi & ACTR10, ACTR1A, CAPZA3, COG4, CTSZ, PPP6C, RAB1B, SEC22C, SEC24C, TMED9 \\
MAPK & DUSP4, EGF, FGF18, FGF20, HSPB1, MAP2K2, MAPKAPK5, RAC1, RAP1A, RASGRP3 \\
Mitochondria Structure & APOOL, APOO, TMEM11, CHCHD6, ATP5ME, MICOS13, ATP5F1C, DNAJC11, DMAC2L, ATP5MF \\
Mitochondrial Transport & ATP5F1A, COA4, COA6, COX17, HSPA9, IDH3G, PITRM1, PMPCA, PMPCB, SLC25A4 \\
mTOR Pathway & CAB39, CAB39L, EIF4EBP1, MLST8, PRKAA2, RPS6KB1, RPTOR, STK11, STRADA, TSC1 \\
Nonsense Mediated Decay & CASC3, EIF4A3, MAGOH, MAGOHB, RBM8A \\
Nuclear Pore & NUP107, NUP133, NUP153, NUP188, NUP205, NUP37, NUP85, NUP93 \\
Nucleolus Structure & FBL, NAT10, NOLC1, NOP58, UTP20 \\
Nucleotide Metabolism & ADSL, ADSS1, ADSS2, ATIC, GMPS, IMPDH1, IMPDH2, PAICS, PFAS, PPAT \\
P53 Stress Signaling & ATM, ATR, CCNG1, CDK1, CHEK1, CHEK2, MDM2, MDM4, TP53, TP73 \\
Pentose Phosphate Pathway & G6PD, TALDO1, DERA, RPE, PGM2, RBKS, PGD, PGLS, RPEL1, PRPS2 \\
Peroxisome Biology & ACOT8, AGPS, BAAT, HMGCL, HSD17B4, MLYCD, PAOX, PEX12, PEX6, PIPOX \\
Prespliceosome Complex & ALYREF, AQR, CRNKL1, DDX5, HNRNPK, LSM2, PLRG1, PRPF4, SMNDC1, SRSF4 \\
Proteasome & PSMA1, PSMA4, PSMB1, PSMB2, PSMB7, PSMA6, PSMA3, PSMB4, PSMA5, PSMB3 \\
Ribosome Large & RPL13A, RPL11, RPL10, RPL23A, RPL30, RPL7A, RPLP2, RPL28, RPL5, RPL27A \\
Ribosome Small & RPS2, RPS6, RPS8, RPS16, RPS11, RPS3A, RPS19, RPS15, RPS4X, RPS9 \\
RNA Polymerase II & POLR2A, POLR2B, POLR2C, POLR2G, POLR2I, POLR2L \\
TCA Cycle & ACO2, DLST, FH, IDH2, IDH3B, MDH2, OGDH, SDHB, SUCLA2, SUCLG2 \\
Tight Junctions & CLDN14, CLDN17, CLDN18, CLDN19, CLDN4, CLDN8, CLDN9, MPP5, PARD6B, PRKCI \\
Translation Initiation Complex & EIF3G, EIF3A, EIF3D, EIF3I, EIF3K, EIF3M, EIF3B, EIF3H, EIF3E, EIF3L \\
Transport Of Fatty Acids & APOD, LCN12, LCN15, LCN9, SLC27A1, SLC27A4, SLC27A6 \\
Tubulin & TUBA3C, TBCC, TBCD, TUBA4B, TUBA8, TUBAL3, TUBA1A, TUBB4B, ARL2, TUBA1B \\
Unfolded Protein Response & CXXC1, DNAJB11, EIF2S3, KHSRP, MBTPS1, SHC1, TATDN2, TLN1, TSPYL2, YIF1A \\
V-ATPase & ATP6V1A, ATP6V, ATP6V1D, ATP6V1E1, ATP6V1F, ATP6V1H \\
\end{tblr}
\label{tab:anax_group_genes}
\tiny
\end{table*}

\end{document}